\documentclass[11pt,twocolumn,twoside]{article}
\usepackage{iccr2024}
\usepackage{siunitx}
\usepackage{tikz}
\usepackage{graphicx}
\usepackage{subcaption}
\usepackage{multirow}
\usepackage{multicol}
\usepackage{algorithm}
\usepackage{algorithmic}

\begin{document}

\title{Iterative Prompt Refinement for Radiation Oncology Symptom Extraction\\Using Teacher-Student Large Language Models} 

\author[1]{Reza Khanmohammadi}
\author[2, 3]{Ahmed I Ghanem}
\author[2]{Kyle Verdecchia}
\author[2]{Ryan Hall}
\author[2]{Mohamed Elshaikh} 
\author[2]{\\ Benjamin Movsas}
\author[2,5,6]{Hassan Bagher-Ebadian}
\author[4,6]{Indrin Chetty}
\author[1]{Mohammad M. Ghassemi\thanks{Shared senior author.}}
\author[2,7]{Kundan Thind*}

\affil[1]{Department of Computer Science and Engineering,
          Michigan State University, East Lansing, USA}

\affil[2]{Department of Radiation Oncology,
          Henry Ford Cancer Institute, Detroit, USA}

\affil[3]{Alexandria Department of Clinical Oncology, Alexandria University, Egypt}

\affil[4]{Department of Radiation Oncology,
          Cedars Sinai Medical Center, Los Angeles, USA}

\affil[5]{Departments of Radiology and Osteopathic Medicine, Michigan State University, East Lansing, USA}

\affil[6]{Department of Physics, Oakland University, Rochester, USA }

\affil[7]{Department of Medicine, Michigan State University, East Lansing, USA}

\maketitle
\thispagestyle{fancy}



\begin{customabstract}
This study introduces a novel teacher-student architecture utilizing Large Language Models (LLMs) to improve prostate cancer radiotherapy symptom extraction from clinical notes. Mixtral, the student model, initially extracts symptoms, followed by GPT-4, the teacher model, which refines prompts based on Mixtral's performance. This iterative process involved 294 single symptom clinical notes across 12 symptoms, with up to 16 rounds of refinement per epoch. Results showed significant improvements in extracting symptoms from both single and multi-symptom notes. For 59 single symptom notes, accuracy increased from 0.51 to 0.71, precision from 0.52 to 0.82, recall from 0.52 to 0.72, and F1 score from 0.49 to 0.73. In 375 multi-symptom notes, accuracy rose from 0.24 to 0.43, precision from 0.6 to 0.76, recall from 0.24 to 0.43, and F1 score from 0.20 to 0.44. These results demonstrate the effectiveness of advanced prompt engineering in LLMs for radiation oncology use. 
\end{customabstract}


\vspace{-5pt}
\section{Introduction}
\vspace{-5pt}
\setlength{\parindent}{20pt}

The development of Large Language Models (LLMs) has marked a transformative era in the field of Artificial Intelligence (AI). These advanced models, built on the foundation of extensive data and sophisticated algorithms, have demonstrated remarkable capabilities in understanding and generating human-like text. In the medical domain, specialized versions of these models such as Med-PaLM 2 \cite{singhal2023expertlevel} and HuatuoGPT \cite{zhang2023huatuogpt} have been tailored to understand and process medical terminology, patient data, and clinical notes. This focus on healthcare has enabled LLMs to assist in diagnostic processes, treatment recommendation, and patient care, thereby revolutionizing the intersection of AI and medicine \cite{karabacak2023embracing}. The ability of these models to handle vast and complex datasets has been pivotal in uncovering insights and augmenting the decision-making process in healthcare \cite{Eghbali}. However, the effectiveness of LLMs in specialized areas like cancer text analysis is often limited by the availability of relevant training data. Since these models learn and evolve based on the data they are exposed to \cite{mardi}, their performance in domains with scarce publicly available data, such as specific cancer-related texts, can be suboptimal \cite{khanmohammadi2023introduction,kamali2023using}. Further, a major challenge for LLMs in specialized domains is the lack of effective prompt engineering, which is essential for enhancing their performance. Prompt engineering is a process that involves iterative refinement of the input and output formats, as well as the incorporation of domain knowledge and terminology. This process can facilitate LLMs to supplement embedded information in specific domains, and thus increase their applicability and effectiveness in niche fields such as oncology \cite{mesko2023prompt}. Recent studies have demonstrated the efficacy of prompt engineering in improving domain-specific inference. Wang et al. \cite{wangp} showed that combining prompt engineering with paraphrases of rare biomedical words can significantly enhance the performance of pre-trained models in the biomedical domain. Similarly, Alsentzer et al. \cite{Alsentzer} utilized a large language model for zero-shot interpretable phenotyping of postpartum hemorrhage from clinical notes, highlighting the model's strong performance in extracting granular concepts associated with the condition. These examples underscore the potential of prompt engineering to refine the capabilities of LLMs in specialized medical tasks, offering a promising avenue for the systematic abstraction of critical concepts in fields such as radiation oncology.

\begin{figure}
    \centering
    \includegraphics[width=1\linewidth]{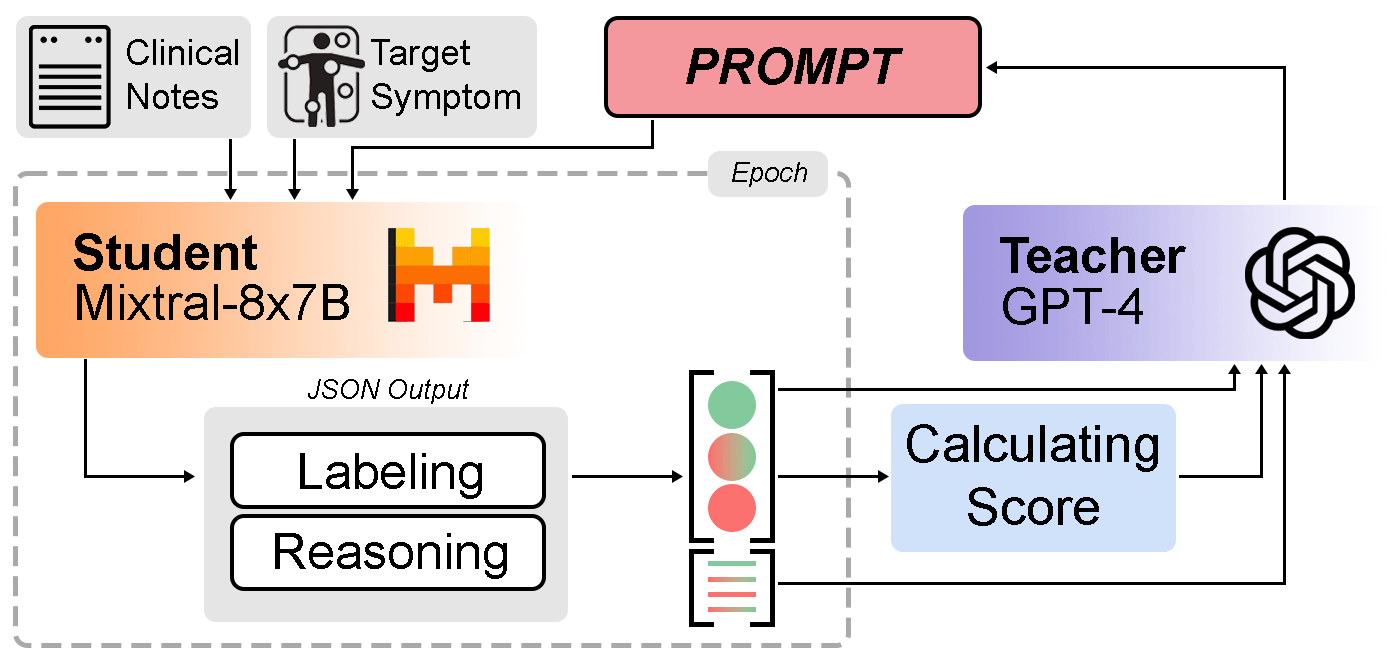}   
    \caption{Schematic representation of our proposed method.}
    \label{fig:framework}
\vspace{-20pt}
\end{figure}
Traditional methodologies in prompt refinement have varied, with some relying on manual tuning by experts, while others explore automated techniques. However, these methods often require a balance between human input, computational resources, and specific task requirements \cite{chen2023unleashing}. This balance is especially crucial in medical text analysis, where precision and context-specific accuracy are paramount. Recent advancements have been observed in the domain of radiation oncology, particularly in the extraction of key concepts from clinical notes, including the identification of toxicity \cite{bittermann, hong}. Consequently, the development of prompt engineering methodologies for the extraction of key concepts in radiation oncology offers the potential to facilitate large-scale, systematic abstraction of critical concepts in this field.

We introduce a novel, automatic framework for prompt refinement, using a teacher-student dynamic between two LLMs: GPT-4 as the teacher and Mixtral as the student. This novel architecture is shown in Figure \ref{fig:framework}. This approach is especially pertinent in the clinical domain for key concept extraction, which is reliant on domain specific knowledge. We demonstrate the application of this architecture with prostate cancer radiotherapy toxicity symptom extraction from clinical notes.

\vspace{-10pt}
\section{Materials and Methods}
\vspace{-5pt}
The architecture employs a novel teacher-student framework designed to iteratively refine the prompts used by LLMs for prostate cancer radiotherapy toxicity symptom extraction
from clinical notes. The student model, Mixtral, is tasked with the initial extraction of symptoms based on provided prompts. The teacher model, GPT-4, subsequently refines these prompts in a series of rounds and epochs based on the student's performance and reasoning.

The student model, Mixtral-8x7B, operates as the primary extractor of symptoms from the clinical notes. Given a prompt and a target symptom, it classifies the symptom as present, negated, or unknown within the notes. The model's output for each note includes the symptom classification and a reasoning for its decision.

The teacher model GPT-4, receives inputs from the student model's performance across all training notes for a given symptom. These inputs include:  \vspace{-5pt}
\begin{itemize}
    \item \textit{Optimal Prompt:} The prompt that has led to the highest level of accuracy in the student's symptom classification. \vspace{-20pt}
    \item \textit{Peak Performance:} The student model's best classification accuracy acquired on the optimal prompt.  \vspace{-5pt}
    \item \textit{Decision-making Reasoning:} Insights into the reasoning behind the student model's classification tied to the optimal prompt. \vspace{-5pt}
    \item \textit{Failed Prompt History:} A list of prompts previously tested within the current epoch, which have yielded inferior results compared to the optimal prompt. \vspace{-5pt}
\end{itemize}
Using these inputs, GPT-4 generates refined prompts intended to improve the student's performance in symptom extraction. This refinement process occurs through a sequence of rounds and epochs.

Full optimization encompasses a series of epochs, each consisting of 16 rounds. In every epoch, the teacher model iteratively refines prompts to improve the student model's symptom extraction performance. At each epoch's start, the teacher is updated with the optimal prompt and peak performance metrics from the previous epoch where it has 16 opportunities (one per round) to refine the prompt and exceed the student's current best performance, measured by accuracy. If, in any of the 16 rounds, the teacher model enhances the student's performance beyond its prior peak, the epoch concludes, and the process advances to the next epoch where the teacher is granted a reset of 16 refinement opportunities. The refinement process terminates if the teacher model fails to improve the student's performance in all 16 rounds of an epoch, signaling the achievement of the student's maximum capability within the framework's parameters.

\begin{table*}[h]
\centering
\footnotesize
\begin{tabular}{l*{6}{S[table-format=1.2]}*{6}{S[table-format=1.2]}*{6}{S[table-format=1.2]}}
\toprule
\multirow{2}{*}{\textbf{Symptom}} & \multicolumn{6}{c}{\textit{\textbf{Single-symptom Notes}}} & \multicolumn{6}{c}{\textit{\textbf{Multi-symptom Notes}}} \\
& \multicolumn{2}{c}{\textbf{Accuracy}} & \multicolumn{2}{c}{\textbf{Precision}} & \multicolumn{2}{c}{\textbf{Recall}} & \multicolumn{2}{c|}{\textbf{F1}} & \multicolumn{2}{c}{\textbf{Accuracy}} & \multicolumn{2}{c}{\textbf{Precision}} & \multicolumn{2}{c}{\textbf{Recall}} & \multicolumn{2}{c}{\textbf{F1}} \\
\cmidrule(lr){2-9} \cmidrule(lr){10-17}
& {Init.} & {Post.} & {Init.} & {Post.} & {Init.} & {Post.} & {Init.} & {Post.} & {Init.} & {Post.} & {Init.} & {Post.} & {Init.} & {Post.} & {Init.} & {Post.} \\
\midrule
Urgency & 0.80 & \textbf{1.00} & 0.87 & \textbf{1.00} & 0.80 & \textbf{1.00} & 0.80 & \textbf{1.00} & 0.37 & \textbf{0.52} & 0.72 & \textbf{0.69} & 0.37 & \textbf{0.52} & 0.27 & \textbf{0.49} \\
Cystitis & 0.40 & \textbf{0.60} & 0.80 & \textbf{0.90} & 0.40 & \textbf{0.60} & 0.53 & \textbf{0.67} & 0.04 & \textbf{0.38} & 0.27 & \textbf{0.78} & 0.04 & \textbf{0.38} & 0.06 & \textbf{0.49} \\
Urinary Obs. & 0.40 & \textbf{0.80} & 0.16 & \textbf{0.67} & 0.40 & \textbf{0.80} & 0.23 & \textbf{0.72} & 0.01 & \textbf{0.12} & \textbf{0.98} & 0.97 & 0.01 & \textbf{0.12} & 0.01 & \textbf{0.19} \\
Dysuria & 0.80 & \textbf{0.80} & 0.65 & \textbf{0.70} & 0.80 & \textbf{0.80} & 0.71 & \textbf{0.73} & 0.56 & \textbf{0.62} & 0.43 & \textbf{0.67} & 0.56 & \textbf{0.62} & 0.44 & \textbf{0.61} \\
Erectile Dys. & \textbf{1.00} & 0.80 & 1.00 & \textbf{1.00} & \textbf{1.00} & 0.80 & \textbf{1.00} & 0.89 & \textbf{0.26} & 0.21 & 0.38 & \textbf{0.62} & \textbf{0.26} & 0.21 & \textbf{0.19} & 0.12 \\
Rectal Bleed. & 0.00 & \textbf{0.20} & 0.00 & \textbf{0.20} & 0.00 & \textbf{0.20} & 0.00 & \textbf{0.20} & 0.16 & \textbf{0.57} & 0.16 & \textbf{0.76} & 0.17 & \textbf{0.57} & 0.10 & \textbf{0.58} \\
Stricture & 0.40 & \textbf{1.00} & 0.16 & \textbf{1.00} & 0.40 & \textbf{1.00} & 0.23 & \textbf{1.00} & 0.04 & \textbf{0.15} & \textbf{0.96} & 0.83 & 0.04 & \textbf{0.15} & 0.02 & \textbf{0.20} \\
Nocturia & \textbf{1.00} & 0.80 & \textbf{1.00} & 0.64 & \textbf{1.00} & 0.80 & \textbf{1.00} & 0.71 & 0.30 & \textbf{0.34} & 0.28 & \textbf{0.75} & 0.30 & \textbf{0.34} & 0.25 & \textbf{0.30} \\
Proctitis & 0.00 & \textbf{0.20} & 0.00 & \textbf{0.80} & 0.00 & \textbf{0.20} & 0.00 & \textbf{0.32} & 0.01 & \textbf{0.20} & 0.74 & \textbf{0.77} & 0.01 & \textbf{0.20} & 0.02 & \textbf{0.32} \\
Hematuria & 0.60 & \textbf{0.60} & 0.80 & \textbf{1.00} & 0.60 & \textbf{0.60} & 0.57 & \textbf{0.70} & 0.72 & \textbf{0.76} & 0.63 & \textbf{0.75} & 0.72 & \textbf{0.76} & 0.66 & \textbf{0.71} \\
Urothelial Carc. & 0.00 & \textbf{1.00} & 0.00 & \textbf{1.00} & 0.00 & \textbf{1.00} & 0.00 & \textbf{1.00} & 0.05 & \textbf{0.70} & 0.99 & \textbf{0.99} & 0.05 & \textbf{0.70} & 0.08 & \textbf{0.81} \\
Incontinence & 0.80 & \textbf{0.80} & 0.87 & \textbf{0.87} & 0.80 & \textbf{0.80} & 0.80 & \textbf{0.80} & 0.39 & \textbf{0.53} & 0.64 & 0.58 & 0.39 & \textbf{0.53} & 0.30 & \textbf{0.51} \\
\midrule
\textbf{Average Score} & 0.51 & \textbf{0.71} & 0.52 & \textbf{0.82} & 0.51 & \textbf{0.71} & 0.49 & \textbf{0.73} & 0.24 & \textbf{0.43} & 0.60 & \textbf{0.76} & 0.24 & \textbf{0.43} & 0.20 & \textbf{0.44} \\ 
\bottomrule
\end{tabular}
\caption{Symptom extraction test results: Initial (Init.) vs Post-refinement (Post)}
\label{tab:toxicity_results}
\vspace{-10pt}
\end{table*}


\begin{algorithm}[t]
\caption{Iterative Prompt Refinement for Symptom Extraction on a target symptom}
\label{alg:prompt_refinement}
\begin{algorithmic}[1]
\REQUIRE Target symptom, Initial prompt, MaxEpochs, MaxRounds
\STATE Initialize Mixtral-8x7B (student model) with initial prompts
\STATE Initialize GPT-4 (teacher model) 
\STATE Set best performance metric to 0
\STATE Set optimal prompt to initial prompt for the symptom
\FOR{epoch = 1 to MaxEpochs}
    \FOR{round = 1 to MaxRounds}
        \STATE Apply current optimal prompt to the student model across all clinical notes
        \STATE Gather labels and reasonings
        \STATE Calculate performance score
        \IF{performance improves over best performance}
            \STATE Update best performance and optimal prompt
            \STATE Break and proceed to the next epoch
        \ENDIF
        \STATE Use the teacher model to refine the optimal prompt based on: \vspace{-10pt}
        \begin{itemize}
            \item Current optimal prompt \vspace{-10pt} 
            \item Best performance metrics \vspace{-10pt}
            \item Reasoning of the student model \vspace{-10pt}
            \item History of failed prompts \vspace{-10pt}
        \end{itemize}
    \ENDFOR
    \IF{no improvement in any of the 16 rounds}
        \STATE Terminate refinement for the symptom
    \ENDIF
\ENDFOR
\end{algorithmic}
\end{algorithm}

The data consisted of clinical notes from a cohort of prostate cancer patients treated with a radiotherapy (RT) dose of 78 Gy between 2013 and 2020. Clinical notes 6 months post-RT were extracted to focus on long term toxicities. 294 notes with presence of single symptoms were selected. Twelve common post-RT toxicity symptoms: Urgency, Cystitis, Urinary Obstruction, Dysuria, Erectile Dysfunction, Rectal Bleeding, Stricture, Nocturia, Proctitis, Hematuria, Urothelial Carcinoma, and Incontinence were selected. For the training set, we utilized 20 notes per symptom, except for Urothelial Carcinoma, where only 15 notes were available. Correspondingly, the test set comprised 5 notes for each symptom, with the exception of Urothelial Carcinoma, which had 4. These notes were labelled with symptom assignment of -1, 0, or 1, signifying its absence, unknown status, or presence, respectively. A separate cohort of 375 clinical notes with multiple symptom within each note were also labelled and used exclusively for validation. Multi-symptom notes ranged from 1 to 10 for the number of distinct symptom mention in each notes, with an average of 5 symptoms mentions per mutli-symptom note. This comprehensive dataset structure and labeling approach provided a robust foundation for evaluating the architecture performance, particularly in demonstrating its scalability and effectiveness in handling complex, multi-symptom clinical notes. Use of clinical notes, and the overall study was reviewed and approved by the institutional review board.

The evaluation methodology utilized accuracy as the primary metric for assessing the student model's performance during the prompt refinement process. To measure the effectiveness of these refined prompts compared to the initial ones, we employed accuracy along with weighted precision, recall, and F1 scores on the test set. These metrics provided a comprehensive view of the refined prompts' performance enhancements over the initial versions. You can find the algorithmic implementation of our method in Algorithm \ref{alg:prompt_refinement}.

\vspace{-10pt}
\section{Results}
\vspace{-5pt}

\begin{figure*}
    \centering
    \includegraphics[width=1\linewidth]{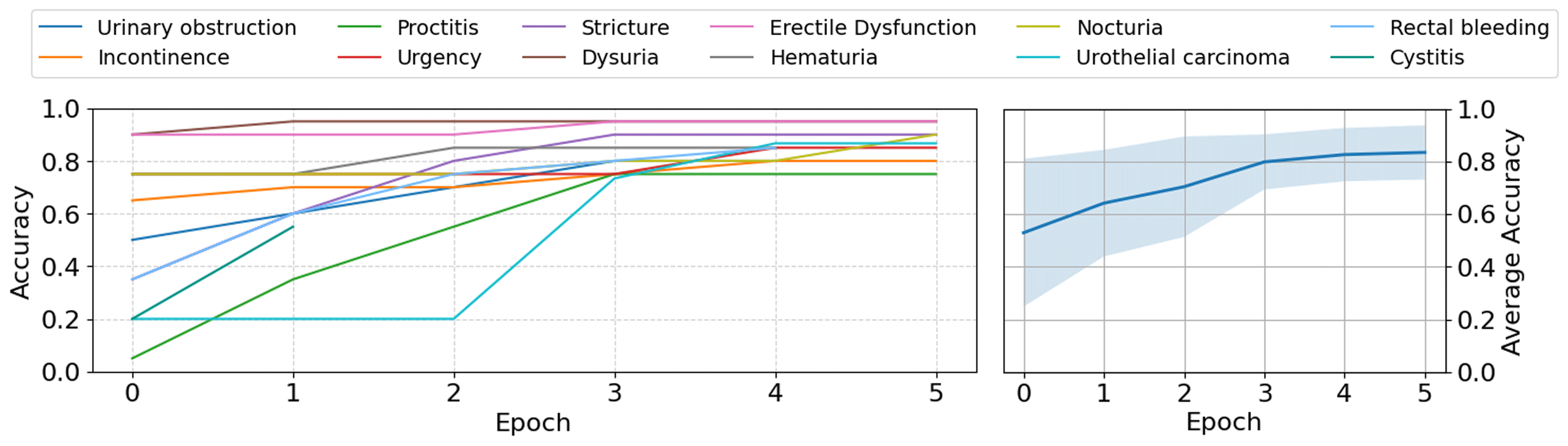}   
    \caption{The evolution of Mixtral's symptom extraction accuracy per different symptoms at different epochs.}
    \label{fig:symp-trends}
\vspace{-10pt}
\end{figure*}

Table \ref{tab:toxicity_results}, summarizes accuracy, precision, recall and F1 scores on both single symptom notes, and multi-symptom notes. Initial and post-refinement refinement results are presented for both categories, with initial performance metrics based on the the initial prompt, and post-refinement performance based on optimized prompt. Improvements across all evaluation metrics is observed. Highlighted entries show performance improvement from initial to post-refined extraction. For single symptom notes, a strong performance increase is observed across all metrics. The average improvement in accuracy from 0.51 to 0.71, precision from 0.52 to 0.82, recall from 0.52 to 0.72, and F1 from 0.49 to 0.73 is observed in single symptom notes. Noteworthy improvements include the symptom "Urgency," where accuracy soared from 0.80 to a perfect 1.00, and precision mirrored this perfect score. Similarly, "Stricture" and "Urothelial Carcinoma" witnessed  accuracy enhancements from 0.40 and 0.00 to 1.00, respectively. Performance improvement across all metrics was observed for multi-symptom notes, with average improvement in accuracy from 0.24 to 0.43, precision from 0.6 to 0.76, recall from 0.24 to 0.43 and F1 from 0.20 to 0.44. The improvement was was not as strong as the single-symptom notes. Notable example of "Rectal Bleeding," showcased an increase in accuracy from 0.16 to 0.57 and an F1 score increase from 0.10 to 0.58.

 Figure 2 depicts the progression of Mixtral's symptom extraction accuracy across different symptoms over successive epochs, as well as the average and standard deviation of symptom accuracy per epoch. A clear trend of improvement is observed in nearly all symptoms as the number of epochs increases, illustrating the efficacy of our prompt refinement process. Notably, symptoms such as "Urothelial Carcinoma" and "Proctitis" show a remarkable upward trajectory, indicating significant gains from the iterative approach. This demonstrates adaptive learning capabilities of this novel architecture.

\vspace{-10pt}
\section{Discussion}
\vspace{-5pt}

This novel student LLM-teacher LLM architecture demonstrates performance improvement in accuracy, precision, recall and F1 scores for prostate cancer RT symptoms. This novel architecture makes contribution in the following key areas: 
\begin{itemize}  
\vspace{-5pt}
\item\textit{\textit{Automatic Prompt Engineering}} This automated method for refining prompts using a teacher-student combination of two LLMs, streamlines the traditionally manual process of prompt optimization.
\vspace{-5pt}
\item\textit{\textit{Zero-Shot Learning}} This framework enhances LLM performance without additional model training, showcasing the efficiency of our zero-shot learning approach.
\vspace{-5pt}
\item\textit{\textit{Local Model Improvement for Privacy}} The improvement of the student model is conducted locally, ensuring sensitive clinical data, like Protected health information (PHI), is processed securely, addressing major data privacy concerns in healthcare.
\end{itemize}

The performance improvement across metrics of accuracy, precision, recall and F1 was observed for both single and multi-symptom notes. The prompt-optimization was performed on single symptom notes, and applied across test set of single symptom notes, and multi-symptom notes. This could potentially have contributed to inferior performance improvement in multi-symptom notes. Limited sample size of single symptom notes utilized for optimization may potentially have resulted in over fitting. Higher sample size for optimization, both single and multi-symptom could potentially further improve robustness, and the performance. Hyperparameter selection for both student and teacher models can also have a significant impact on the overall optimization process. Repetition studies were performed to tune hyperparameters of student model to ensure absolute consistency in response over three repetitions. As such, low \texttt{temperature} (0.2) and \texttt{top\_p} (0.1) were utilized for the student model, and high \texttt{temperature} (2) and \texttt{top\_p} (0.9) were used for the teacher model in this architecture. This may have limitations and further experimentation may lead to better results.  

\vspace{-10pt}
\section{Conclusion}
\vspace{-5pt}
In conclusion, this novel architecture presents a pathway for self-optimized local LLM agents that can extract key concepts in medical notes with a zero shot learning approach. The complete local inference by the student model is suitable for protecting privacy of healthcare data. This system provides proof-of-concept in the research setting, and is promising for the potential of natural language processing methods to support clinical care.

\printbibliography

@article{bittermann,
    title = "An End-to-End Natural Language Processing System for Automatically Extracting Radiation Therapy Events From Clinical Texts",
      author = "Bitterman DS Goldner E Finan S Harris D, Durbin EB Hochheiser H Warner JL Mak RH, Miller T Savova GK.",
      journal = "Int J Radiat Oncol Biol Phys.",
      year = "2023",
      volume = "117",
      number = "3",
      pages = "262-273",
      doi = "10.1016/j.ijrobp.2023.03.055",
      pmid = "36990288",
      pmcid = "PMC10522797"   
}

@article{hong,
    author = {Hong, Julian C and Fairchild, Andrew T and Tanksley, Jarred P and Palta, Manisha and Tenenbaum, Jessica D},
    title = "{Natural language processing for abstraction of cancer treatment toxicities: accuracy versus human experts}",
    journal = {JAMIA Open},
    volume = {3},
    number = {4},
    pages = {513-517},
    year = {2020},
    month = {12},
    issn = {2574-2531},
    doi = {10.1093/jamiaopen/ooaa064},
    url = {https://doi.org/10.1093/jamiaopen/ooaa064},
    eprint = {https://academic.oup.com/jamiaopen/article-pdf/3/4/513/36625805/ooaa064.pdf},
}

@misc{singhal2023expertlevel,
      title={Towards Expert-Level Medical Question Answering with Large Language Models}, 
      author={Karan Singhal and Tao Tu and Juraj Gottweis and Rory Sayres and Ellery Wulczyn and Le Hou and Kevin Clark and Stephen Pfohl and Heather Cole-Lewis and Darlene Neal and Mike Schaekermann and Amy Wang and Mohamed Amin and Sami Lachgar and Philip Mansfield and Sushant Prakash and Bradley Green and Ewa Dominowska and Blaise Aguera y Arcas and Nenad Tomasev and Yun Liu and Renee Wong and Christopher Semturs and S. Sara Mahdavi and Joelle Barral and Dale Webster and Greg S. Corrado and Yossi Matias and Shekoofeh Azizi and Alan Karthikesalingam and Vivek Natarajan},
      year={2023},
      eprint={2305.09617},
      archivePrefix={arXiv},
      primaryClass={cs.CL}
}

@misc{zhang2023huatuogpt,
      title={HuatuoGPT, towards Taming Language Model to Be a Doctor}, 
      author={Hongbo Zhang and Junying Chen and Feng Jiang and Fei Yu and Zhihong Chen and Jianquan Li and Guiming Chen and Xiangbo Wu and Zhiyi Zhang and Qingying Xiao and Xiang Wan and Benyou Wang and Haizhou Li},
      year={2023},
      eprint={2305.15075},
      archivePrefix={arXiv},
      primaryClass={cs.CL}
}

@article{karabacak2023embracing,
  title = {Embracing Large Language Models for Medical Applications: Opportunities and Challenges},
  author = {Karabacak, Mert and Margetis, Konstantinos},
  journal = {Cureus},
  year = {2023},
  volume = {15},
  issue = {5},
  pages = {e39305},
  doi = {10.7759/cureus.39305},
  url = {https://doi.org/10.7759/cureus.39305},
  issn = {2168-8184 (Print), 2168-8184 (Electronic)},
  pmid = {37378099},
}

@article{mesko2023prompt,
  author = {Meskó, Bertalan},
  title = {Prompt Engineering as an Important Emerging Skill for Medical Professionals: Tutorial},
  journal = {Journal of Medical Internet Research},
  year = {2023},
  volume = {25},
  pages = {e50638},
  doi = {10.2196/50638},
  url = {https://doi.org/10.2196/50638},
  issn = {1439-4456 (Print), 1438-8871 (Electronic)},
}

@misc{chen2023unleashing,
      title={Unleashing the potential of prompt engineering in Large Language Models: a comprehensive review}, 
      author={Banghao Chen and Zhaofeng Zhang and Nicolas Langrené and Shengxin Zhu},
      year={2023},
      eprint={2310.14735},
      archivePrefix={arXiv},
      primaryClass={cs.CL}
}

@article {Alsentzer,
	author = {Emily Alsentzer and Matthew J Rasmussen and Romy Fontoura and Alexis L Cull and Brett Beaulieu-Jones and Kathryn J Gray and David W Bates and Vesela P Kovacheva},
	title = {Zero-shot Interpretable Phenotyping of Postpartum Hemorrhage Using Large Language Models},
	elocation-id = {2023.05.31.23290753},
	year = {2023},
	doi = {10.1101/2023.05.31.23290753},
	publisher = {Cold Spring Harbor Laboratory Press},
	eprint = {https://www.medrxiv.org/content/early/2023/06/01/2023.05.31.23290753.full.pdf},
	journal = {medRxiv}
}

@misc{kamali2023using,
      title={Using Persuasive Writing Strategies to Explain and Detect Health Misinformation}, 
      author={Danial Kamali and Joseph Romain and Huiyi Liu and Wei Peng and Jingbo Meng and Parisa Kordjamshidi},
      year={2023},
      eprint={2211.05985},
      archivePrefix={arXiv},
      primaryClass={cs.CL}
}

@article{mardi,
    author = {Mardikoraem, Mehrsa and Wang, Zirui and Pascual, Nathaniel and Woldring, Daniel},
    title = "{Generative models for protein sequence modeling: recent advances and future directions}",
    journal = {Briefings in Bioinformatics},
    volume = {24},
    number = {6},
    pages = {bbad358},
    year = {2023},
    month = {10},
    issn = {1477-4054},
    doi = {10.1093/bib/bbad358},
    url = {https://doi.org/10.1093/bib/bbad358},
    eprint = {https://academic.oup.com/bib/article-pdf/24/6/bbad358/52312609/bbad358.pdf},
}

@misc{khanmohammadi2023introduction,
      title={An Introduction to Natural Language Processing Techniques and Framework for Clinical Implementation in Radiation Oncology}, 
      author={Reza Khanmohammadi and Mohammad M. Ghassemi and Kyle Verdecchia and Ahmed I. Ghanem and Luo Bing and Indrin J. Chetty and Hassan Bagher-Ebadian and Farzan Siddiqui and Mohamed Elshaikh and Benjamin Movsas and Kundan Thind},
      year={2023},
      eprint={2311.02205},
      archivePrefix={arXiv},
      primaryClass={cs.CL}
}

@inproceedings{wangp,
    title = "Prompt Combines Paraphrase: Teaching Pre-trained Models to Understand Rare Biomedical Words",
    author = "Wang, Haochun  and
      Liu, Chi  and
      Xi, Nuwa  and
      Zhao, Sendong  and
      Ju, Meizhi  and
      Zhang, Shiwei  and
      Zhang, Ziheng  and
      Zheng, Yefeng  and
      Qin, Bing  and
      Liu, Ting",
    editor = "Calzolari, Nicoletta  and
      Huang, Chu-Ren  and
      Kim, Hansaem  and
      Pustejovsky, James  and
      Wanner, Leo  and
      Choi, Key-Sun  and
      Ryu, Pum-Mo  and
      Chen, Hsin-Hsi  and
      Donatelli, Lucia  and
      Ji, Heng  and
      Kurohashi, Sadao  and
      Paggio, Patrizia  and
      Xue, Nianwen  and
      Kim, Seokhwan  and
      Hahm, Younggyun  and
      He, Zhong  and
      Lee, Tony Kyungil  and
      Santus, Enrico  and
      Bond, Francis  and
      Na, Seung-Hoon",
    booktitle = "Proceedings of the 29th International Conference on Computational Linguistics",
    month = oct,
    year = "2022",
    address = "Gyeongju, Republic of Korea",
    publisher = "International Committee on Computational Linguistics",
    url = "https://aclanthology.org/2022.coling-1.122",
    pages = "1422--1431"
}

@article{Eghbali,
author = {Eghbali, Niloufar and Siegal, Daniel and Klochko, Chad and Ghassemi, Mohammad},
year = {2023},
month = {06},
pages = {118-127},
title = {Automation of Protocoling Advanced MSK Examinations Using Natural Language Processing Techniques},
volume = {2023},
journal = {AMIA Joint Summits on Translational Science proceedings. AMIA Joint Summits on Translational Science}
}

\end{document}